\documentclass[11pt]{article}

\usepackage{array}
\newcommand{\PreserveBackslash}[1]{\let\temp=\\#1\let\\=\temp}
\newcolumntype{C}[1]{>{\PreserveBackslash\centering}p{#1}}
\newcolumntype{R}[1]{>{\PreserveBackslash\raggedleft}p{#1}}
\newcolumntype{L}[1]{>{\PreserveBackslash\raggedright}p{#1}}
\usepackage[]{acl}

\usepackage{paralist}

\usepackage{times}
\usepackage{latexsym}
\usepackage{amssymb}
\usepackage{hyperref}
\usepackage{xurl}
\usepackage{wrapfig}
\usepackage{times}
\usepackage{tipa}
\usepackage{latexsym}
\usepackage{microtype}
\usepackage{graphicx}
\usepackage{adjustbox}
\usepackage{multirow}
\usepackage[inline]{enumitem}
\usepackage{graphicx}
\usepackage{url}
\usepackage{textcomp}
\usepackage{amssymb}
\usepackage{caption}
\usepackage{subcaption}
\usepackage{color,soul}
\usepackage{paralist}
\usepackage{float}
\usepackage{placeins}
\usepackage{aliascnt}
\usepackage{mathtools}
\usepackage{amsmath} 
\usepackage{hhline}
\usepackage{booktabs}
\usepackage{footnote}
\usepackage{soul}
\usepackage{multirow}
\usepackage[normalem]{ulem}
\usepackage{amsmath,amssymb,stmaryrd} 
\usepackage{amssymb}
\usepackage{pifont}
\useunder{\uline}{\ul}{}
\usepackage{arydshln}
\newcommand\blfootnote[1]{%
  \begingroup
  \renewcommand\thefootnote{}\footnote{#1}%
  \addtocounter{footnote}{-1}%
  \endgroup
}

\usepackage[T1]{fontenc}
\usepackage[utf8]{inputenc}

\usepackage{listings}

\lstdefinestyle{mystyle}{
    backgroundcolor=\color{gray!10},   
    commentstyle=\color{green},
    keywordstyle=\color{blue},
    numberstyle=\tiny\color{gray},
    stringstyle=\color{red},
    basicstyle=\ttfamily\footnotesize,
    breakatwhitespace=false,         
    breaklines=true,                 
    captionpos=b,                    
    keepspaces=true,                 
    numbers=left,                    
    numbersep=5pt,                  
    showspaces=false,                
    showstringspaces=false,
    showtabs=false,                  
    tabsize=4
}

\usepackage{microtype}

\usepackage{inconsolata}
\usepackage{enumitem} 
\usepackage{graphicx}

\title{Assessing Thai Dialect Performance in LLMs with Automatic Benchmarks and Human Evaluation}

\author{
Peerat Limkonchotiwat\textsuperscript{$\spadesuit$*}, 
Wuttikorn Ponwitayarat\textsuperscript{$\heartsuit$*},
Lalita Lowphansirikul\textsuperscript{$\heartsuit$}, \\
\textbf{Potsawee Manakul}\textsuperscript{$\clubsuit$}, 
\textbf{Can Udomcharoenchaikit}\textsuperscript{$\heartsuit$},
\textbf{Ekapol Chuangsuwanich}\textsuperscript{$\diamondsuit$}, \\
\textbf{Sarana Nutanong}\textsuperscript{$\heartsuit$}\\
  \textsuperscript{$\spadesuit$}AI Singapore, Singapore, \textsuperscript{$\clubsuit$}SCB 10X, Thailand\\
  \textsuperscript{$\heartsuit$}School of Information Science and Technology, VISTEC, Thailand,\\
  \textsuperscript{$\diamondsuit$}Department of Computer Engineering, Chulalongkorn University, Thailand \\
  \texttt{peerat@aisingapore.org}, \texttt{wuttikorn.p\_s22@vistec.ac.th}
  }

\author{
 \textbf{Peerat Limkonchotiwat\textsuperscript{1,4,*}},
 \textbf{Kanruethai Masuk,\textsuperscript{2,*}},
 \textbf{Surapon Nonesung\textsuperscript{3,*}},
\\
\textbf{Chalermpun Mai-On\textsuperscript{2,*}},
 \textbf{Sarana Nutanong\textsuperscript{2}},
 \textbf{Wuttikorn Ponwitayarat\textsuperscript{2,*}},
 \\
 \textbf{Potsawee Manakul\textsuperscript{3,*}},
\\
 \textsuperscript{1}AI Singapore,
 \textsuperscript{2}Vidyasirimedhi Institute of Science and Technology, 
 \\
 \textsuperscript{3}SCB10X,
 \textsuperscript{4}National University of Singapore,
\\
 \small{
   \textbf{Correspondence:} \href{mailto:peerat@aisingapore.org}{peerat@aisingapore.org}
 }
}

\begin{document}
\maketitle
\blfootnote{\textsuperscript{*}Equal contributions}
\begin{abstract}
Large language models show promising results in various NLP tasks.
Despite these successes, the robustness and consistency of LLMs in underrepresented languages remain largely unexplored, especially concerning local dialects. 
Existing benchmarks also focus on main dialects, neglecting LLMs' ability on local dialect texts.  
In this paper, we introduce a Thai local dialect benchmark covering Northern (Lanna), Northeastern (Isan), and Southern (Dambro) Thai, evaluating LLMs on five NLP tasks: summarization, question answering, translation, conversation, and food-related tasks. 
Furthermore, we propose a human evaluation guideline and metric for Thai local dialects to assess generation fluency and dialect-specific accuracy.
Results show that LLM performance declines significantly in local Thai dialects compared to standard Thai, with only proprietary models like GPT-4o and Gemini2 demonstrating some fluency\footnote{Data and Evaluation Tool: \url{https://github.com/mrpeerat/Thai_local_benchmark}}.

\end{abstract}

\section{Introduction}
Large language models (LLMs) play a crucial role in natural language processing (NLP) by significantly enhancing downstream performance across diverse tasks and languages. 
Studies have shown that LLMs surpass traditional multilingual models~\cite{devlin-etal-2019-bert,conneau-etal-2020-unsupervised} in tasks such as question answering and summarization~\cite{touvron2023llamaopenefficientfoundation,pipatanakul2023typhoonthailargelanguage}.
One key reason is that LLMs have significantly larger parameter counts and training datasets, often exceeding traditional models by a factor of 26, from 0.3 billion to 8 billion parameters. 
As a result, LLMs demonstrate greater robustness than traditional models across most scenarios, particularly in low-resource languages such as Thai~\cite{DBLP:conf/acl/PengpunUBL24,phatthiyaphaibun-etal-2024-chie}, Indonesian~\cite{cahyawijaya-etal-2024-cendol,cahyawijaya-etal-2023-nusacrowd}, and Tagalog~\cite{montalan2024kalahihandcraftedgrassrootscultural,gamboa-lee-2025-filipino}.

\begin{figure*}[t!]
    \hspace*{-5mm}
    \includegraphics[width=1.05\textwidth]{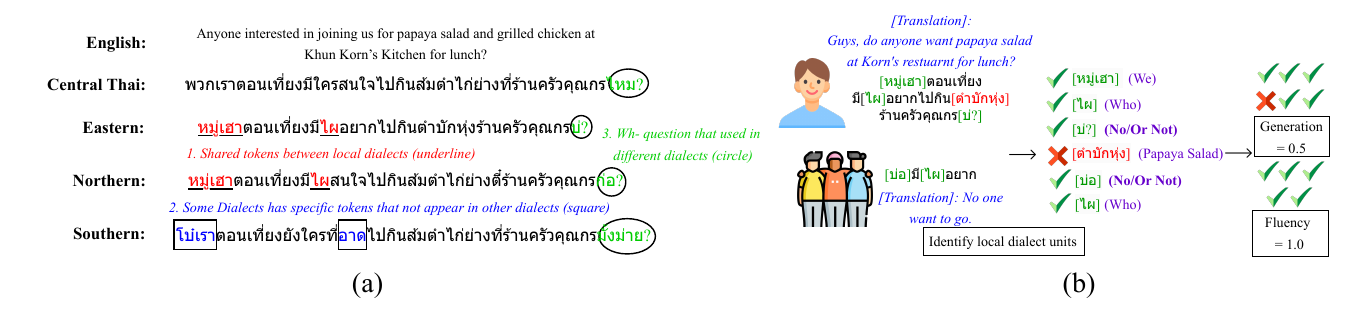}
    \vspace{-9mm}
    \caption{(a) We showed some unique characteristics of Thai local dialects compared to Central Thai: 1. shared tokens in local dialects; 2. unique words for each local dialect; and 3. WH- question tokens that are different in each local dialect. (b) Illustration of our local dialect evaluation metric. The example demonstrates the assessment of generation and fluency, highlighting variations in spelling and pronunciation among different Thai dialects.}
    \label{fig:main_text}
\vspace{-5mm}
\end{figure*}

Researchers are increasingly evaluating LLMs in underrepresented languages, particularly Thai, to assess their robustness and consistency in low-resource languages. 
Thai-H6~\cite{kim-etal-2025-representing} introduces an evaluation benchmark for assessing LLMs' understanding of Thai cultural knowledge. 
SEACrowd~\cite{lovenia-etal-2024-seacrowd} introduces a Southeast Asian benchmark for evaluating LLMs on local and cross-lingual texts, audios, and images, focusing on natural language generation and understanding.
The Thai LLM benchmark\footnote{\url{https://huggingface.co/spaces/ThaiLLM-Leaderboard/leaderboard}} extends SEACrowd by incorporating LLM-as-a-Judge and Thai Exam datasets for a Thai language evaluation. 
However, these benchmarks primarily focus on the Central Thai dialect. 
Therefore, challenges observed in the Central dialect may also reflect underlying issues in local dialects.

Although there are 69 million Thai speakers, only about 20 million have Central Thai as their first language.
The rest primarily grow up speaking other dialects, including approximately 15.2 million who speak Northeastern Thai (Isan), 6 million who speak Northern Thai (Lanna), and 4.5 million who speak Southern Thai (Dambro) or other local varieties.
%
%
As shown in Figure~\ref{fig:main_text}a, a distinct feature of Thai local dialects is the blending of words between Central Thai and local dialects, as well as among different local dialects. 
Moreover, each local dialect has distinct words, some unique to that dialect and others shared with other dialects but not found in others.
%
%
Recent studies examine the evaluation of fluency and understanding in local dialects of Indonesian~\cite{10.4108/eai.15-9-2021.2315616,cahyawijaya-etal-2023-nusacrowd,Septianingtias31122024} and Filipino~\cite{tabajunda-2018-linguistic,guevara-etal-2024-philippine}.
Evaluating the fluency and understanding of LLMs in Thai local dialects remains underexplored.

In this paper, we introduce a benchmark for evaluating local Thai dialects--Isan, Lanna, and Dambro--to assess LLMs' understanding of these dialects. 
Specifically, we evaluate LLMs on five tasks -- summarization, question answering, translation, conversation, and food -- using instructions, inputs, and labels written in local dialects. 
Additionally, we introduce human evaluation alongside a new \emph{Thai local dialect guideline and metric} to assess LLMs' capabilities in understanding and generating local dialects. 
The new metric evaluates the generation and fluency of local dialects, rewarding or penalizing the model for fluency in a manner consistent with human speakers.
Our experimental results show that LLM performance significantly declines on summarization and translation tasks when inputs are written in local dialects, compared to inputs in Central Thai. 
Furthermore, only proprietary models--GPT and Gemini--are capable of speaking local Thai dialects, as shown by both human evaluation and our new metric.

Our contributions are as follows: We introduce a benchmark for evaluating three Thai local dialects. 
We also propose a guideline and metric to assess the generation and fluency of LLMs in Thai local dialects. 
Lastly, we are the first to investigate Thai local dialects in the context of LLMs, presenting comprehensive evaluations using both traditional evaluation metrics and human assessment.

\section{Benchmark and Metric}

\subsection{Benchmark Formulation}

\noindent
\textbf{Data Selection.} We create a new benchmark for evaluating three Thai local dialects--Isan, Lanna, and Dambro--drawing on examples from the Thai LLM leaderboard to assess model performance. 
We select generation tasks—Question Answering (QA), summarization, and translation—because they are more complex and better suited for assessing model knowledge than more straightforward multiple-choice QA tasks. 
We randomly selected 20 samples for each task, resulting in 80 samples per dialect, where the translation task was both English to local dialects and vice versa.
In addition, we include 10 topics on food (i.e., \emph{asking for a food recipe of [FOOD\_NAME]}) and conversations between two individuals discussing Thai traditional and cultural topics (i.e., \emph{talking bout [TOPIC\_NAME]}). 
The full topics can be found in Appendix~\ref{subsec:topics}.

\noindent
\textbf{Gold Standard.}
Our dataset uses the gold standard provided by the original Central Thai examples.
Note that we do not create labels for the food and conversation topics, as these tasks are open-ended and lack a definitive correct answer. 
Instead, we rely on human evaluation to assess these tasks. 
The number of samples is 100, all written in Central Thai.

\noindent
\textbf{Translation.} We enlisted native speakers of each local dialect to translate the 100 samples from Central Thai into their respective dialects. 
This includes translating the input, context (for QA and summarization tasks), prompts, and labels, except for those related to food and conversation samples. 
Furthermore, all annotators are asked to review translations completed by their peers within the same region to ensure quality and consistency. 
Therefore, the total number of samples is 400, comparable to previous LLM-as-a-Judge and Human evaluation benchmarks~\cite{zheng2023judging,qin-etal-2024-infobench}.
For \emph{the full information} of data statistics, prompts, topics, annotator demographics, and guidelines, see Appendix~\ref{sec:data_stat}.

\begin{table*}[h!]
\centering
\setlength\doublerulesep{0.5pt}

\scalebox{0.65}{
\setlength\columnsep{23pt}

\begin{tabular}{l|l|c|c|c|c|c|c|c|c}
\hline
\textbf{Model} & \textbf{Dialect} & \textbf{QA} & \textbf{$\Delta$Diff}                & \textbf{Summarization} & \textbf{$\Delta$Diff}               & \textbf{Translation (E-L)} & \textbf{$\Delta$Diff}               & \textbf{Translation (L-E)} & \textbf{$\Delta$Diff}               \\ \hline \hline
Llama-3.1-8b   & Central            & 60.74       & -                             & 20.07                  & -                            & 11.41                             & -                            & 44.61                             & -                            \\ 
Typhoon1.5-8b  & Central            & 33.36       & -                             & 11.52                  & -                            & 5.14                              & -                            & 47.35                             & -                            \\ 
Llama-3.1-70b  & Central            & 56.44       & -                             & 23.03                  & -                            & 18.11                             & -                            & 51.07                             & -                            \\ 
Typhoon1.5-70b & Central            & 46.99       & -                             & 10.83                  & -                            & 8.46                              & -                            & 50.54                             & -                            \\ 
GPT-4o & Central            & 56.03       & -                             & 14.67                  & -                            & 21.926                              & -                            & 62.82                             & -                            \\ 
Gemini2 & Central            & 53.73       & -                             & 17.70                  & -                            & 20.76                              & -                            & 58.10                             & -                            \\ \hline
Llama-3.1-8b   & Isan              & 58.84       & {\color[HTML]{FF0000} 1.90}    & 16.47                  & {\color[HTML]{FF0000} 3.60}   & 3.17                              & {\color[HTML]{FF0000} 8.24}  & 6.59                              & {\color[HTML]{FF0000} 38.02} \\ 
Typhoon1.5-8b  & Isan              & 28.48       & {\color[HTML]{FF0000} 4.88}   & 8.54                   & {\color[HTML]{FF0000} 2.98}  & 0.83                              & {\color[HTML]{FF0000} 4.31}  & 31.6                              & {\color[HTML]{FF0000} 15.75} \\ 
Llama-3.1-70b  & Isan              & 57.59       & {\color[HTML]{34A853} ↑1.15}  & 18.77                  & {\color[HTML]{FF0000} 4.26}  & 8.98                              & {\color[HTML]{FF0000} 9.13}  & 33.74                             & {\color[HTML]{FF0000} 17.33} \\ 
Typhoon1.5-70b & Isan              & 45.32       & {\color[HTML]{FF0000} 1.67}   & 10.87                  & {\color[HTML]{34A853} ↑0.04} & 2.82                              & {\color[HTML]{FF0000} 5.64}  & 41.81                             & {\color[HTML]{FF0000} 8.73}  \\ 
GPT-4o & Isan              & 44.21       & {\color[HTML]{FF0000} 11.82}   & 10.39                  & {\color[HTML]{FF0000} 4.28} & 14.29                              & {\color[HTML]{FF0000} 7.63}  & 55.26                             & {\color[HTML]{FF0000} 7.56}  \\
Gemini2 & Isan              & 40.58       & {\color[HTML]{FF0000} 13.15}   & 12.68                  & {\color[HTML]{FF0000} 5.02} & 13.31                              & {\color[HTML]{FF0000} 7.45}  & 51.74                             & {\color[HTML]{FF0000} 6.36}  \\ \hline
Llama-3.1-8b   & Lanna             & 55.66       & {\color[HTML]{FF0000} 5.08}   & 18.05                  & {\color[HTML]{FF0000} 2.02}  & 2.46                              & {\color[HTML]{FF0000} 8.95}  & 7.27                              & {\color[HTML]{FF0000} 37.34} \\ 
Typhoon1.5-8b  & Lanna             & 24.39       & {\color[HTML]{FF0000} 8.97}   & 7.28                   & {\color[HTML]{FF0000} 4.24}  & 0.49                              & {\color[HTML]{FF0000} 4.65}  & 18.48                             & {\color[HTML]{FF0000} 28.87} \\ 
Llama-3.1-70b  & Lanna             & 50.97       & {\color[HTML]{FF0000} 5.47}   & 17.41                  & {\color[HTML]{FF0000} 5.62}  & 3.73                              & {\color[HTML]{FF0000} 14.38} & 15.92                             & {\color[HTML]{FF0000} 35.15} \\ 
Typhoon1.5-70b & Lanna             & 44.74       & {\color[HTML]{FF0000} 2.25}   & 10.17                  & {\color[HTML]{FF0000} 0.66}  & 4.76                              & {\color[HTML]{FF0000} 3.70}   & 39.01                             & {\color[HTML]{FF0000} 11.53} \\ 
GPT-4o & Lanna             & 45.73       & {\color[HTML]{FF0000} 10.30}   & 9.78                  & {\color[HTML]{FF0000} 4.89}  & 12.02                              & {\color[HTML]{FF0000} 9.90}   & 54.83                             & {\color[HTML]{FF0000} 7.99} \\ 
Gemini2 & Lanna             & 40.86       & {\color[HTML]{FF0000} 12.87}   & 11.70                  & {\color[HTML]{FF0000} 6.0}  & 7.68                              & {\color[HTML]{FF0000} 13.08}   & 47.28                             & {\color[HTML]{FF0000} 10.82} \\ \hline
Llama-3.1-8b   & Dambro             & 61.38       & {\color[HTML]{34A853} ↑0.64}  & 16.91                  & {\color[HTML]{FF0000} 3.16}  & 7.22                              & {\color[HTML]{FF0000} 4.19}  & 24.02                             & {\color[HTML]{FF0000} 20.59} \\ 
Typhoon1.5-8b  & Dambro             & 31.34       & {\color[HTML]{FF0000} 2.02}   & 9.18                   & {\color[HTML]{FF0000} 2.34}  & 4.20                               & {\color[HTML]{FF0000} 0.94}  & 27.81                             & {\color[HTML]{FF0000} 19.54} \\ 
Llama-3.1-70b  & Dambro             & 66.96       & {\color[HTML]{34A853} ↑10.52} & 17.2                   & {\color[HTML]{FF0000} 5.83}  & 7.66                              & {\color[HTML]{FF0000} 10.45} & 27.10                              & {\color[HTML]{FF0000} 23.97} \\ 
Typhoon1.5-70b & Dambro             & 52.10        & {\color[HTML]{34A853} ↑5.11}  & 9.74                   & {\color[HTML]{FF0000} 1.09}  & 0.78                              & {\color[HTML]{FF0000} 7.68}  & 39.19                             & {\color[HTML]{FF0000} 11.35} \\ 
GPT-4o  & Dambro             & 49.28       & {\color[HTML]{FF0000} 6.75}   & 12.87                   & {\color[HTML]{FF0000} 1.80}  & 12.95                               & {\color[HTML]{FF0000} 8.97}  & 45.32                             & {\color[HTML]{FF0000} 17.50} \\ 
Gemini2  & Dambro             & 47.91       & {\color[HTML]{FF0000} 5.82}   & 12.07                   & {\color[HTML]{FF0000} 5.63}  & 9.13                               & {\color[HTML]{FF0000} 11.63}  & 23.46                             & {\color[HTML]{FF0000} 34.64} \\ \hline \hline
\end{tabular}}
\vspace{-2mm}
\caption{ The main results of our benchmark. We calculate the $\Delta$Diff by comparing Central and local dialects at the same model. In addition, E equals to English and L equals to local dialects.}
\vspace{-5mm}
\label{tab:main_results}
\end{table*}

\subsection{Human Evaluation for Local Dialects} \label{subsec:new_metric}
Traditional evaluation metrics, such as BLEU~\cite{papineni-etal-2002-bleu} and ROUGE-L~\cite{lin-2004-rouge}, struggle to assess dialectal text accurately.  
A key challenge is their reliance on tokenization, which fails when dialect words are missing from standard dictionaries, resulting in inaccurate segmentation. 
Furthermore, the absence of a standardized writing system leads to multiple valid spellings for the same word. 
Therefore, these traditional metrics that compare generated text to a fixed reference fail to account for such variations, often misclassifying correct responses as errors. 
These limitations underscore the urgency of developing more flexible evaluation methods that reflect the linguistic diversity of dialects.

To bridge these gaps, we conduct our local dialect evaluation metric as follows.
\begin{compactenum}[1]
\item Annotators evaluate whether the generated output includes accurate local units, covering all forms of writing: nouns, phrases, and sentences.
\item Different dialect spellings are allowed if they share the same pronunciation. We ensure this by transliterating the word and comparing its phonetic transcriptions.
\item We divide our assessment into two categories: language \emph{Generation} (similar to recall) and language \emph{Fluency} (similar to precision). First, annotators (native local dialect speakers) evaluate language \emph{Generation} by assigning a score of 0 (no local dialect generated), 0.5 (partially generated), or 1 (fully generated). Next, annotators assess the correctness and naturalness (\emph{Fluency}) of the generated dialectal words using the same scoring system as \emph{Generation}: 0 (completely incorrect), 0.5 (partially correct), and 1 (fully correct and natural). The final score is calculated by the average over two annotators.
\end{compactenum} 
Figure \ref{fig:main_text}(b) illustrates an example of our evaluation using the proposed metrics. Some extracted units have the same meaning but are written differently. Both writings are considered correct because they share the same pronunciation.
In addition, the averages of annotator agreements (Appendix~\ref{appendix:agreement}) from three dialects are 0.7969 and 0.7449 for \emph{Generation} and \emph{Fluency}, respectively. 
We discussed the possibility of replacing humans with LLMs in Appendix~\ref{subsec:llm-as-a-judge}.

\section{Experimental Setups and Results}

\subsection{Setup}
To evaluate Thai local dialects, we use well-known Thai and multilingual LLMs for 8 and 70 billion parameters, namely Typhoon1.5~~\footnote{\url{https://huggingface.co/collections/scb10x/typhoon-15x-6648d7b07ab33d141d6648b6}} and Llama3.1~\footnote{\url{https://huggingface.co/collections/meta-llama/llama-31-669fc079a0c406a149a5738f}}.
We also evaluate proprietary LLMs, such as GPT-4o and Gemini2, in our benchmark.
We use BLEU for translation and ROUGE-L for QA and summarization according to the original benchmark~\cite{lovenia-etal-2024-seacrowd}, including the generation setting for a fair comparison with previous benchmark works.

\begin{table*}[h!]
\renewcommand{\arraystretch}{1.2}

\centering
\scalebox{0.65}{
\setlength{\tabcolsep}{8pt}
\setlength\columnsep{10pt}
\begin{tabular}{l|C{2cm}ccc|cccc|cccc}
\hline
\multicolumn{1}{c|}{\multirow{2}{*}{\textbf{A vs B}}}                   & \multicolumn{4}{c|}{\textbf{Isan}}                                                                                                                                & \multicolumn{4}{c|}{\textbf{Lanna}}                                                                                                                                & \multicolumn{4}{c}{\textbf{Dambro}}                                                                                                                               \\ \cline{2-13} 
\multicolumn{1}{c|}{}                                                   & \multicolumn{1}{l|}{\textbf{A}} & \multicolumn{1}{l|}{\textbf{B}} & \multicolumn{1}{l|}{\textbf{Both}} & \multicolumn{1}{l|}{\textbf{None}} & \multicolumn{1}{l|}{\textbf{A}} & \multicolumn{1}{l|}{\textbf{B}} & \multicolumn{1}{l|}{\textbf{Both}} & \multicolumn{1}{l|}{\textbf{None}} & \multicolumn{1}{|l|}{\textbf{A}} & \multicolumn{1}{l|}{\textbf{B}} & \multicolumn{1}{l|}{\textbf{Both}} & \multicolumn{1}{l}{\textbf{None}} \\ \hline \hline
\begin{tabular}[c]{@{}l@{}}Typhoon1.5-8b vs Llama3.1-8b\end{tabular}   & \multicolumn{1}{c|}{1}                 & \multicolumn{1}{c|}{-}                 & \multicolumn{1}{c|}{-}                  & \textbf{19}                              & \multicolumn{1}{c|}{-}                 & \multicolumn{1}{c|}{-}                 & \multicolumn{1}{c|}{-}                  & \textbf{20}                              & \multicolumn{1}{c|}{1}                 & \multicolumn{1}{c|}{-}                 & \multicolumn{1}{c|}{-}                  & \textbf{19}                              \\ \hline
\begin{tabular}[c]{@{}l@{}}Typhoon1.5-70b vs Llama3.1-70b\end{tabular} & \multicolumn{1}{c|}{-}                 & \multicolumn{1}{c|}{1}                 & \multicolumn{1}{c|}{-}                  & \textbf{19}                              & \multicolumn{1}{c|}{2}                 & \multicolumn{1}{c|}{-}                 & \multicolumn{1}{c|}{-}                  & \textbf{18}                              & \multicolumn{1}{c|}{4}                 & \multicolumn{1}{c|}{-}                 & \multicolumn{1}{c|}{-}                  & \textbf{16}                              \\ \hline
\begin{tabular}[c]{@{}l@{}}Gemini2 vs Typhoon1.5-70b\end{tabular}      & \multicolumn{1}{c|}{\textbf{20}}       & \multicolumn{1}{c|}{-}                 & \multicolumn{1}{c|}{-}                  & -                                        & \multicolumn{1}{c|}{\textbf{19}}       & \multicolumn{1}{c|}{-}                 & \multicolumn{1}{c|}{-}                  & 1                                        & \multicolumn{1}{c|}{\textbf{16}}       & \multicolumn{1}{c|}{2}                 & \multicolumn{1}{c|}{1}                  & 1                                        \\ \hline
\begin{tabular}[c]{@{}l@{}}Gemini2 vs GPT-4o\end{tabular}            & \multicolumn{1}{c|}{4}                 & \multicolumn{1}{c|}{3}                 & \multicolumn{1}{c|}{\textbf{13}}        & -                                        & \multicolumn{1}{c|}{\textbf{18}}       & \multicolumn{1}{c|}{-}                 & \multicolumn{1}{c|}{2}                  & -                                        & \multicolumn{1}{c|}{\textbf{16}}       & \multicolumn{1}{c|}{2}                 & \multicolumn{1}{c|}{2}                  & -                                        \\ \hline \hline
\end{tabular}}
\vspace{-2mm}
\caption{ Human fluency preference on conversation and food topics. The full details are described in Section~\ref{subsec:human_eval}.}
\vspace{-5mm}
\label{tab:human_eval}
\end{table*}

\subsection{Traditional Metric Results} \label{subsec:main_results}
Table~\ref{tab:main_results} shows a significant decline in translation performance for both 8B and 70B models, with the largest drop observed in Lanna. 
Results in Isan and Dambro were more mixed. 
In particular, QA performance improved in both Dambro and Isan.
One possible explanation is that ROUGE-L measures only text overlap, assessing correctness based on matching words rather than fluency. 
Appendix~\ref{appendix:output_examples} provides LLMs' outputs, revealing that ROUGE-L assigned high scores even when Llama-3.1-70B failed to produce local dialects. 
In particular, \emph{none of the models produced responses in Thai local dialects}, even when explicitly prompted to do so.
If reference labels are not written in local dialects, models may still achieve high scores despite failing to generate dialectal text, as shown in Table~\ref{subsec:word_overlap} that QA and summarization exhibit higher word overlap in Central and local dialects than translation.
This highlights the need for alternative evaluation techniques that assess fluency rather than relying solely on exact match metrics.

\subsection{Human Evaluation} \label{subsec:human_eval}
To verify that \emph{existing models do not generate Thai local dialects} and \emph{the traditional metric problem in Thai local dialects}, we conduct a human evaluation focused on fluency in food and conversation topics.
We also recruit three native speakers per dialect to assess the fluency of LLM outputs, selecting one of four options: (A) for Prefer A, (B) for Prefer B, (Both) for liking both models equally, or (None) for disliking without any preference. 
In addition, we average the answer from three annotators to formulate the final answer.
The annotator guideline can be found in Appendix~\ref{subsec:guideline}. 

Table~\ref{tab:human_eval} confirms that all open-source models fail to generate Thai local dialects, as expected. 
Typhoon and Llama received the highest number of \emph{both are the worst} ratings from annotators. 
For the Dambro dialect, Typhoon demonstrates some ability to generate local dialects, aligning with Table~\ref{tab:main_results}, where it showed improvement in Dambro texts.
Gemini generates local dialects more fluently and outperforms other models in nearly every comparison. 
However, these results reflect only a preference-based evaluation, while the actual fluency of LLMs in local dialects remains unmeasured.
This underscores the need for an alternative metric to assess dialect fluency beyond exact match scores or subjective human ratings.

\subsection{Our Local Dialect Metric}
To better assess LLMs' fluency and generative ability in local dialects, we evaluate them using our proposed metric. 
As described in Section~\ref{subsec:new_metric}, two native speakers evaluate Typhoon1.5-70B and Gemini2 for QA, summarization, and translation.

Table~\ref{tab:gam_metric} shows that when we focus only on evaluating the generation and fluency performance in Thai local dialects, only Gemini2 speaks Thai local dialects.
Although Typhoon1.5-70b outperformed Gemini2 in the traditional metric performance in Table~\ref{tab:main_results}, we found a significant gap between these models.
Gemini2 achieves over $\sim$92.58 points in the Lanna dialect and $\sim$84.49 points in Dambro, while Typhoon achieves less than 31 points and 12 points in Dambro and Lanna, respectively.
Moreover, the results of our experiment also conform with Table~\ref{tab:human_eval} that Gemini2 is better than Typhoon1.5 with more deep analysis results, the Generation and Fluency score.
%
%
This emphasizes the robustness of our metric, which can measure the ability to speak the local dialect fluently, unlike the traditional metrics (Table~\ref{tab:main_results}) and human preference (Table~\ref{tab:human_eval}).  

\begin{table}[h!]
\vspace{-3mm}

\centering
\setlength\doublerulesep{0.5pt}
\scalebox{0.65}{
\setlength{\tabcolsep}{3pt}
\begin{tabular}{L{1.45cm}|ccc|ccc|ccc}
\hline
\multicolumn{1}{c|}{\multirow{2}{*}{\textbf{Model}}} &
  \multicolumn{3}{c|}{\textbf{Isan}} &
  \multicolumn{3}{c|}{\textbf{Lanna}} &
  \multicolumn{3}{c}{\textbf{Dambro}} \\ \cline{2-10} 
\multicolumn{1}{c|}{} &
  \multicolumn{1}{c|}{\textbf{G.}} &
  \multicolumn{1}{c|}{\textbf{F.}} &
  \textbf{Avg.} &
  \multicolumn{1}{c|}{\textbf{G.}} &
  \multicolumn{1}{c|}{\textbf{F.}} &
  \textbf{Avg.} &
  \multicolumn{1}{c|}{\textbf{G.}} &
  \multicolumn{1}{c|}{\textbf{F.}} &
  \textbf{Avg.} \\ \hline \hline
Typhoon&
  \multicolumn{1}{c|}{10.34} &
  \multicolumn{1}{c|}{15.74} &
  13.04 &
  \multicolumn{1}{c|}{10.71} &
  \multicolumn{1}{c|}{13.22} &
  11.97 &
  \multicolumn{1}{c|}{30.50} &
  \multicolumn{1}{c|}{30.00} &
  30.25 \\ 
Gemini2 &
  \multicolumn{1}{c|}{\textbf{86.57}} &
  \multicolumn{1}{c|}{\textbf{82.41}} &
  \textbf{84.49} &
  \multicolumn{1}{c|}{\textbf{91.33}} &
  \multicolumn{1}{c|}{\textbf{93.99}} &
  \textbf{92.58} &
  \multicolumn{1}{c|}{\textbf{75.50}} &
  \multicolumn{1}{c|}{\textbf{68.50}} &
  \textbf{71.83} \\ \hline \hline
\end{tabular}}
\vspace{-2mm}
\caption{ The results from our propose metric. Note that G. is Generation and F. is Fluency.}
\vspace{-6mm}
\label{tab:gam_metric}
\end{table}

\section{Conclusion}
Our study highlights the significant performance gap in LLMs when processing Thai local dialects.
While proprietary models like GPT-4 and Gemini demonstrate some capability in understanding and generating local dialects, most LLMs struggle with fluency and accuracy. 
Our proposed benchmark, human evaluation, and a new guideline and metrics provide valuable insights into these limitations, paving the way for future improvements in multilingual and dialect-aware LLMs.

\section*{Limitation}
\begin{itemize}
    
    \item We select only the main local dialects in Thailand since some dialects do not have a writing system, only speaking.
    \item We experiment only with top-performance Thai LLMs, such as Typhoon1.5, Llama3.1, Gemini2, and GPT-4o. We acknowledge other Thai LLMs, such as WangchanLION and OpenThaiGPT. However, we found that only the typhoon model has 8 and 70 billion model parameters, covering all experiments in our paper.
    \item Extending the proposed metric to other languages. We acknowledge that our paper does not explain extending our metric to other languages. Our proposed metric can extend to other languages when human annotators who can speak local dialects are available. We will study this in our future work. 
    \item Limited number of annotators. We use only five annotators for each local dialect. However, the agreement score demonstrated a high agreement in the annotation part.  
\end{itemize}

\section*{Ethical Statement}
All annotators are volunteers. 
In addition, we demonstrate the annotator details in Appendix~\ref{subsec:annotators}. 
Moreover, we will release our Thai local dialect and human evaluation results, for both the training and our human evaluation methods, to the public with the original license of each dataset, such as XL-SUM~\cite{hasan-etal-2021-xl} and iAPP~\cite{kobkrit_viriyayudhakorn_2021_4539916}.
In addition, we have the dataset that was created by hand, and we will release it with license CC-BY-SA, similar to previous works. 

\section*{Acknowledgement}

This research is supported by the National Research Foundation, Singapore under its National Large Language Models Funding Initiative. Any opinions, findings and conclusions or recommendations expressed in this material are those of the author(s) and do not reflect the views of National Research Foundation, Singapore.
We also would like to thank all the annotators in this work for their volunteer work. 

\bibliography{acl_latex}

\appendix

\section{Data Statistics and Annotators}
\label{sec:data_stat}

\subsection{Data Statistics}
As shown in Table~\ref{tab:data_stat}, we have a total of 100 samples for each local dialect.
We use summarization, QA, and translation for the traditional metric, BLEU, and ROUGE-L. In contrast, we use conversation and food topics for human evaluation since these topics do not have the perfect answer, as in previous tasks.
In addition, the topics of food and conversation are culturally relevant, such as local food in Southern or Northern Thailand, and the conversation topic is events and culture in each part of Thailand.
In addition, we have 40 samples for translation since it is forward and backward translation. 
Moreover, the QA samples are from iAPP~~\cite{kobkrit_viriyayudhakorn_2021_4539916}, the summarization samples are from XL-SUM~\cite{hasan-etal-2021-xl}, and the translation, food, and conversation samples are formulated by annotators. 

\begin{table}[h!]

\centering
\setlength\doublerulesep{0.5pt}
\scalebox{0.58}{
\setlength{\tabcolsep}{3pt}
\begin{tabular}{|l|c|c|c|c|c|}
\hline
\multicolumn{1}{|c|}{\textbf{Dialects}} &
  \textbf{\begin{tabular}[c]{@{}c@{}}\#Sample of \\ Summarization\end{tabular}} &
  \textbf{\begin{tabular}[c]{@{}c@{}}\#Sample \\ of QA\end{tabular}} &
  \textbf{\begin{tabular}[c]{@{}c@{}}\#Sample \\ of Transaltion\end{tabular}} &
  \textbf{\begin{tabular}[c]{@{}c@{}}\#Sample \\ of Conversation\end{tabular}} &
  \textbf{\begin{tabular}[c]{@{}c@{}}\#Sample \\ of Food\end{tabular}} \\ \hline
Isan  & 20 & 20 & 40 & 10 & 10 \\ \hline
Lanna & 20 & 20 & 40 & 10 & 10 \\ \hline
Dambro  & 20 & 20 & 40 & 10 & 10 \\ \hline
\end{tabular}}
\caption{ Data Statistical: Number of Samples}
\vspace{-4mm}
\label{tab:data_stat}
\end{table}

\subsection{Word Overlap} \label{subsec:word_overlap}
As shown in Table~\ref{tab:word_overlap}, we observe that the translation had the least overlap compared to other tasks.
This conforms with the results in Table~\ref{tab:main_results} that the least word overlap yields the lowest penalty, directly affecting the downstream task performance.
However, the word-overlap score still cannot explain the fluency performance of LLMs in speaking Thai local dialects.
\begin{table}[h!]

\centering
\setlength\doublerulesep{0.5pt}
\scalebox{0.6}{
\setlength{\tabcolsep}{3pt}
\begin{tabular}{|l|c|c|c|}
\hline
\multicolumn{1}{|c|}{\textbf{Dialects}} & \textbf{Summarization} & \textbf{QA} & \textbf{Translation} \\ \hline
Isan  & 0.9628 & 0.9167  & 0.9067 \\ \hline
Lanna & 0.9767 & 0.93318 & 0.9072 \\ \hline
Dambro  & 0.9598 & 0.9008  & 0.9036 \\ \hline
\end{tabular}}
\caption{ Data Statistical: Word Overlap}
\vspace{-4mm}
\label{tab:word_overlap}
\end{table}

\subsection{Annotators} \label{subsec:annotators}

We list the author's biographic as follows.
For Lanna, we have two women and three men, and annotators are between 25 and 30 years old.
For Isan, we have three women and two men; the annotators are between 25 and 30 years old and 31 and 35 years old.
For Dambro, we have three women and two men, and annotators are between 25 and 30 years old.
All annotators are born in non-Central Thailand and speak local and Central dialects.

\subsection{Annotator Agreement}
\label{appendix:agreement}
In this study, we calculate the annotator agreement of our proposed metric using Cohen's Kappa score. 
We use two annotators in this experiment, and both of them need to annotate the same sample (100\% sample overlap).
As shown in Table~\ref{tab:agreement}, we found that Isan and Dambro have a similar agreement score, ~0.7 points, with a substantial agreement.
Moreover, we found almost perfect agreement on Lanna. 
This is because the model can speak Lanna better than other dialects. 
When the annotators assign the score, it is easier for them to judge it.

\begin{table}[h!]

\centering
\setlength\doublerulesep{0.5pt}
\scalebox{0.58}{
\setlength{\tabcolsep}{3pt}
\begin{tabular}{|c|cc|cc|cc|}
\hline
\multirow{2}{*}{\textbf{Agreement}} & \multicolumn{2}{c|}{\textbf{Lanna}}                         & \multicolumn{2}{c|}{\textbf{Isan}}                          & \multicolumn{2}{c|}{\textbf{Dambro}}                         \\ \cline{2-7} 
                                    & \multicolumn{1}{c|}{\textbf{Generation}} & \textbf{Fluency} & \multicolumn{1}{c|}{\textbf{Generation}} & \textbf{Fluency} & \multicolumn{1}{c|}{\textbf{Generation}} & \textbf{Fluency} \\ \hline
\multicolumn{1}{|l|}{Cohen\_kappa}  & \multicolumn{1}{c|}{0.9315}              & 0.7319           & \multicolumn{1}{c|}{0.7078}              & 0.7077           & \multicolumn{1}{c|}{0.7514}              & 0.7951           \\ \hline
\end{tabular}}
\caption{ Data Statistical: Annotator agreement of our proposed metric.}
\vspace{-4mm}
\label{tab:agreement}
\end{table}

\subsection{Example Code}
Since we use all the generation setup from \citet{lovenia-etal-2024-seacrowd}, we can run all the experiments with one line code, similar to SEACrowd.
For example:
\lstdefinestyle{mystyle}{
    backgroundcolor=\color{gray!10},   
    commentstyle=\color{green},
    keywordstyle=\color{blue},
    numberstyle=\tiny\color{gray},
    stringstyle=\color{red},
    basicstyle=\ttfamily\footnotesize,
    breakatwhitespace=false,         
    breaklines=true,                 
    captionpos=b,                    
    keepspaces=true,                 
    numbers=left,                    
    numbersep=5pt,                  
    showspaces=false,                
    showstringspaces=false,
    showtabs=false,                  
    tabsize=4
}

\lstset{style=mystyle}

\begin{lstlisting}[language=bash, caption=Example Code]
#!/bin/bash
MODEL_NAME='meta-llama/Llama-3.1-8B-Instruct'
python evaluation/main_local_prompt_batch.py south ${MODEL_NAME} 0 1
python evaluation/main_local_prompt_batch.py north ${MODEL_NAME} 0 1
\end{lstlisting}
In addition, the output will be in the CSV format, including scores according to tasks.
The benchmark can be run on a single A100 or H100 (800 GB) within < 1 hour. 

\subsection{Guideline for Annotators} \label{subsec:guideline}
\noindent
\textbf{Translation.} From Section~\ref{subsec:main_results}, we translate from Central Thai to Thai local dialects using the following guideline:

\begin{quote}
Translation Task

In this task, we need your help to revise the translation of the prompt and its response from English to your native language.

Comparing Original and translated texts and then editing the translation to be more human (write your revision in the edit column)

The goal is to make the translation results look more like human writing

Note that the edit columns can’t be null. You need to edit all the translations.

Do not remove or edit emojis, hashtags, or special characters 
If the text represents gender (she/he), please change it to the general context (I, you, they, them)

All texts should be edited since it is not 100\% natural. There is no empty field in the revision column.

\end{quote}

\noindent
\textbf{Model Comparison Guideline.} From Section~\ref{subsec:human_eval}, we ask three annotators to compare model A vs. B where we mean the answer to get the final answer of each sample.
Annotators are blinded to model identities for all examples.
In addition, we average the answer from three annotators into the final answer.

\begin{quote}
Guidelines for Evaluating Model A and Model B

Evaluation Objective:
Assess the performance of Model A compared to Model B in responding to provided questions/instructions/prompts.

Evaluate based on the following two aspects: 1. Accuracy of local dialect/language usage 2. Accuracy of the response

Evaluation Options: You must choose only one option for each criterion:

(i) A is better than B: Model A performs better than Model B in the given aspect

(ii) B is better than A: Model B performs better than Model A in the given aspect

(iii) A and B are equally good: Both models perform equally well in the given aspect

(iv) A and B are equally poor: Both models perform equally poorly in the given aspect

Evaluation Justification:
Provide a short reason explaining why you chose that score (repeating reasons is allowed if the outputs are similar).

\end{quote}

\subsection{Food and Conversation Topics} \label{subsec:topics}
\begin{figure*}[t!]
    \includegraphics[width=0.5\textwidth]{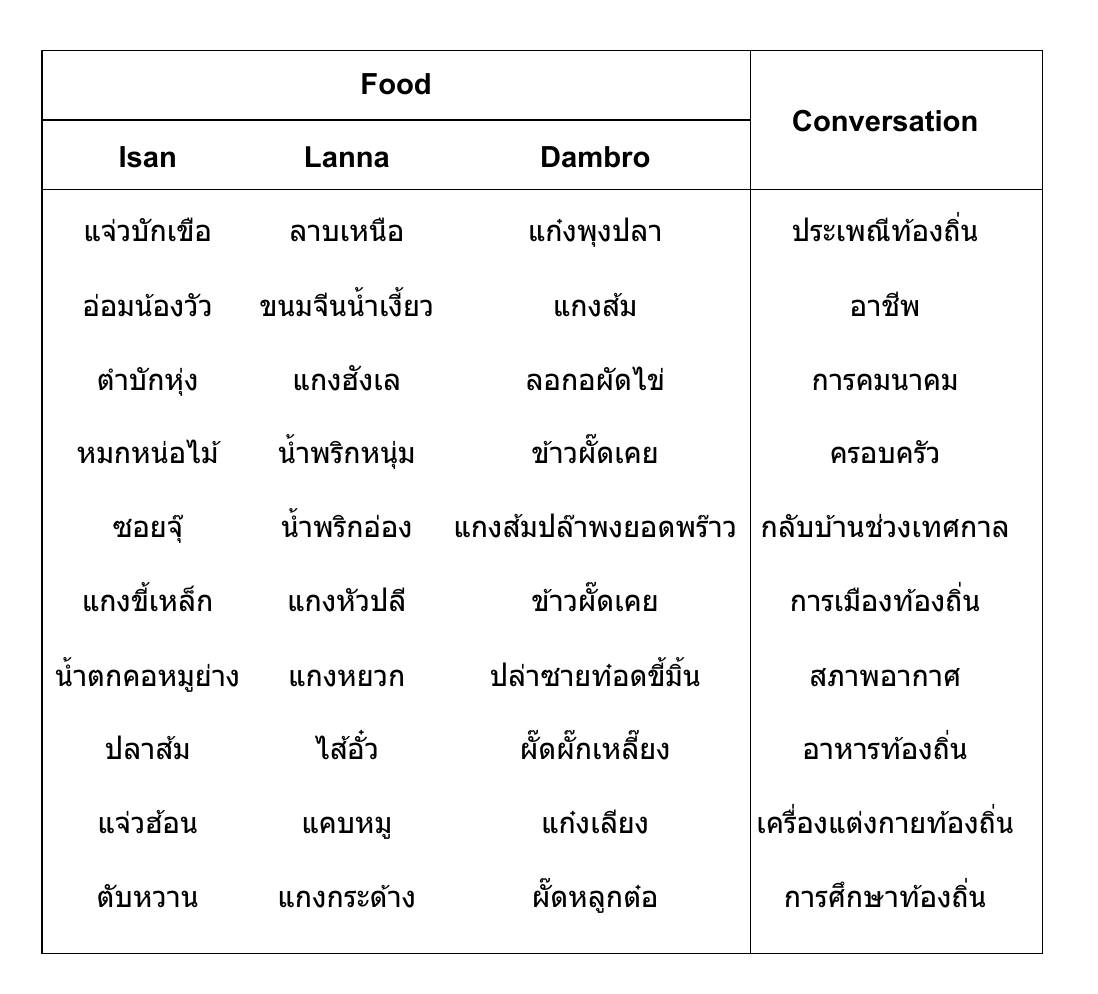}
    \centering
    \vspace{-3mm}
    \caption{Food and Conversation topics. Note that these food names are very local in each part of Thailand.}
    \label{fig:food_convo_topics}
\vspace{-3mm}
\end{figure*}

We provide the full food and conversation topics in Figure~\ref{fig:food_convo_topics}. 

\subsection{Prompts}
\begin{figure*}[t!]
    \includegraphics[width=1\textwidth]{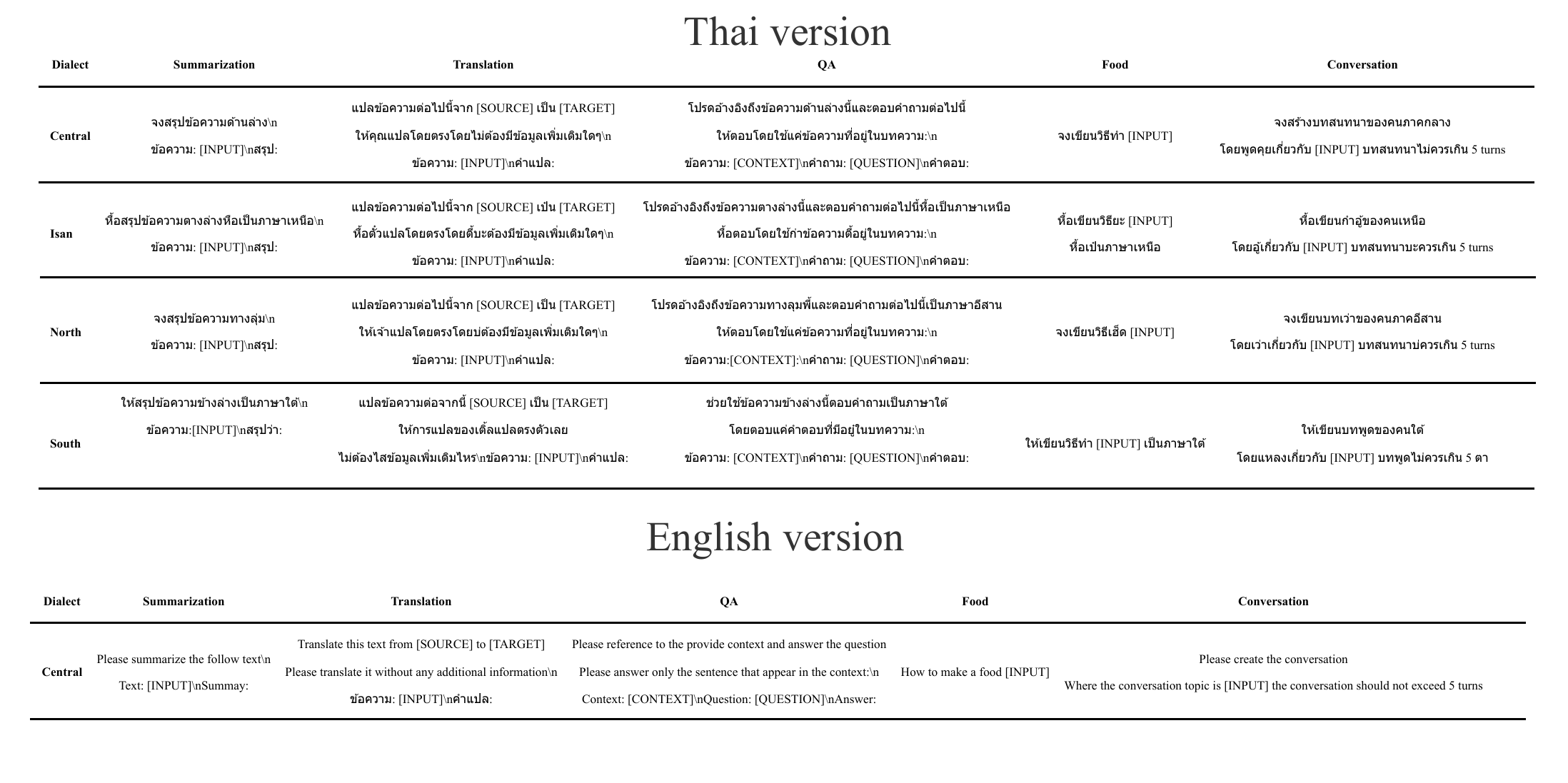}
    \centering
    \vspace{-8mm}
    \caption{The Thai prompts that we used in our experiments with the translation version.}
    \label{fig:prompts}
\vspace{-3mm}
\end{figure*}

We demonstrate the prompt we used in our experiment in Figure~\ref{fig:prompts}.
Note that all prompts are written in local dialects, and we also added instructions for performing in local dialects.

\begin{figure*}[t!]
    \includegraphics[width=1\textwidth]{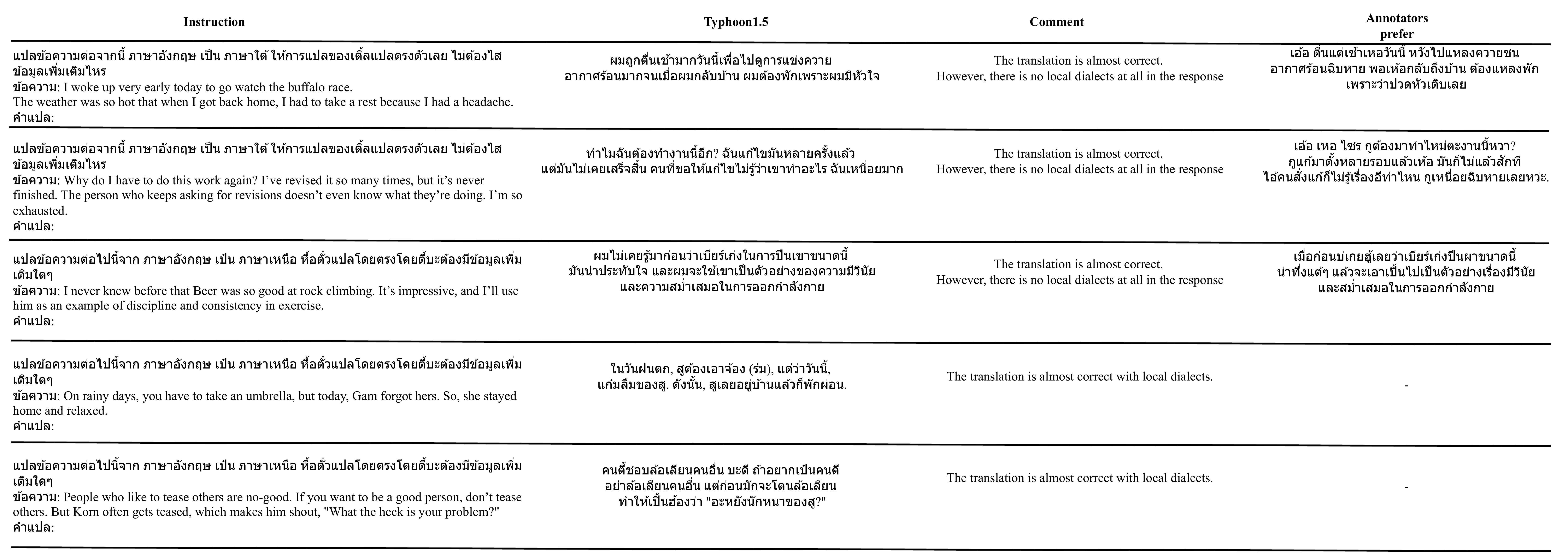}
    \centering
    \vspace{-8mm}
    \caption{Example from the translation task.}
    \label{fig:example}
\vspace{-3mm}
\end{figure*}

\section{LLM-as-a-judge} \label{subsec:llm-as-a-judge}
We acknowledge that the limitation of our proposed metric is relying on human 100\%.
We use native speakers to judge the fluency and generation score for LLMs' output.
However, we tried to use LLM-as-a-judge to solve this problem and found that LLMs (i.e., Gemini2 and GPT4o) are not good enough to understand Thai local dialects fluency, resulting in inaccurate judgment compared to native speakers.
Although Gemini2 achieves a high score in Table~\ref{tab:gam_metric}, when it acts like a judge, it fails to give a similar fluency score to the human, where the preliminary results show that the correlation between Geimini2 and Humans on Lanna is only 53.6 points (Spearman's correlation). 
Therefore, we omit the LLM-as-a-judge from our work.

\section{Output Examples} \label{appendix:output_examples}
In this study, we show the incorrect and correct examples when we focus on the local dialect output.
We divided it into two categories: (i) answer correctly with no local dialects, and (ii) answer correctly with local dialects to make it easier to understand local Thai dialects.
Note that we use the output from Typhoon1.5-70b and Gemini2.

As shown in Figure~\ref{fig:example}, we notice that while Typhoon can translate the text correctly, there are no local dialects in this case.
In contrast, Gemini2 (anotator prefer) can translate from English to local dialect correctly. 
Moreover, we also observe Figure that both models code-switch between Central and local dialects, as underscored in Figure~\ref{fig:main_text}. 
This suggests that the gap in making LLMs speak Thai local dialects is significantly broad and needs more attention.

\end{document}